\documentclass[dvipsnames,sigconf,balance=false]{aamas} 
\usepackage{tikz,tikzscale,svg}
\usepackage{pgfplots}
\usepackage[font=small]{caption}
\usepackage{subcaption}
\usepackage{array}
\usepackage[normalem]{ulem}

\DeclareUnicodeCharacter{2212}{−}
\usepgfplotslibrary{groupplots,dateplot}
\usetikzlibrary{patterns,shapes.arrows,matrix}
\pgfplotsset{compat=newest}

\acmSubmissionID{574}

\title[Heterogeneous Multi-Robot Reinforcement Learning]{Heterogeneous Multi-Robot Reinforcement Learning}
\author{Matteo Bettini}
\affiliation{
  \institution{University of Cambridge}
  \city{Cambridge}
  \country{United Kingdom}}
\email{mb2389@cl.cam.ac.uk}

\author{Ajay Shankar}
\affiliation{
  \institution{University of Cambridge}
  \city{Cambridge}
  \country{United Kingdom}}
\email{as3233@cl.cam.ac.uk}

\author{Amanda Prorok}
\affiliation{
  \institution{University of Cambridge}
  \city{Cambridge}
  \country{United Kingdom}}
\email{asp45@cl.cam.ac.uk}

\begin{abstract}
Cooperative multi-robot tasks can benefit from heterogeneity in the robots' physical and behavioral traits. In spite of this, traditional Multi-Agent Reinforcement Learning (MARL) frameworks lack the ability to explicitly accommodate policy heterogeneity, and typically constrain agents to share neural network parameters. This enforced homogeneity limits application in cases where the tasks benefit from heterogeneous behaviors. In this paper, we crystallize the role of heterogeneity in MARL policies. Towards this end, we introduce Heterogeneous Graph Neural Network Proximal Policy Optimization (HetGPPO), a paradigm for training heterogeneous MARL policies that leverages a Graph Neural Network for differentiable inter-agent communication. HetGPPO allows communicating agents to learn heterogeneous behaviors while enabling fully decentralized training in partially observable environments. We complement this with a taxonomical overview that exposes more heterogeneity classes than previously identified. To motivate the need for our model, we present a characterization of techniques that homogeneous models can leverage to emulate heterogeneous behavior, and show how this “apparent heterogeneity’’ is brittle in real-world conditions. Through simulations and real-world experiments, we show that: (i) when homogeneous methods fail due to strong heterogeneous requirements, HetGPPO succeeds, and, (ii) when homogeneous methods are able to learn apparently heterogeneous behaviors, HetGPPO achieves higher resilience to both training and deployment noise.
\end{abstract}


\newcommand{\BibTeX}{\rm B\kern-.05em{\sc i\kern-.025em b}\kern-.08em\TeX}

\newcommand{\R}{\mathbb{R}}

\begin{document}

%%% The following commands remove the headers in your paper. For final 
%%% papers, these will be inserted during the pagination process.

\pagestyle{fancy}
\fancyhead{}

%%% The next command prints the information defined in the preamble.

\maketitle 

%%%%%%%%%%%%%%%%%%%%%%%%%%%%%%%%%%%%%%%%%%%%%%%%%%%%%%%%%%%%%%%%%%%%%%%%%%%%%%%%
\section{Introduction}
\label{sec:intro}

% Motivation
Multi-robot systems deployed to tackle complex cooperative tasks can often benefit from heterogeneous physical and/or behavioral traits to fulfill their mission.
%Robot teams comprised of multiple agents not only enable a greater force in numbers (more robots for the same task), but also in diversity (robots with different capabilities for the same task).
Such heterogeneous systems have been leveraged in applications such as disaster response~\cite{michael2014collaborative}, collaborative mapping~\cite{boroson20193d}, agriculture~\cite{ju2019modeling}, and package transport~\cite{gerkey2002pusher}.
However, synthesizing optimal decentralized policies for these tasks can be computationally hard, and typically scales exponentially with the number of agents~\cite{bernstein2002complexity}. 
While faster and scalable solutions exist, such as metaheuristics~\cite{braysy2005vehicle}, they lack in optimality.
Multi-Agent Reinforcement Learning (MARL)~\cite{zhang2021multi} can be used as a scalable approach to find near-optimal solutions to these problems.
However, MARL algorithms without inter-agent communication cannot be easily applied to real-world robotic problems, where partial observability of individual agents is pervasive.
%Communication is key to overcoming the natural partial observability of individual agents (situated in the real world), and to enabling cooperation.
Communication is key to overcoming this
partial observability,
and to enable cooperation.
Our work deals with \textit{heterogeneous multi-robot reinforcement learning}, a paradigm located at the boundary of MARL (with inter-agent communication) and heterogeneous multi-robot systems.

%Among MARL paradigms, we focus on those that leverage inter-agent communication to overcome partial observability of individual agents and enable cooperation. 

\begin{figure}[t]
    \centering 
    \includegraphics[width=\linewidth]{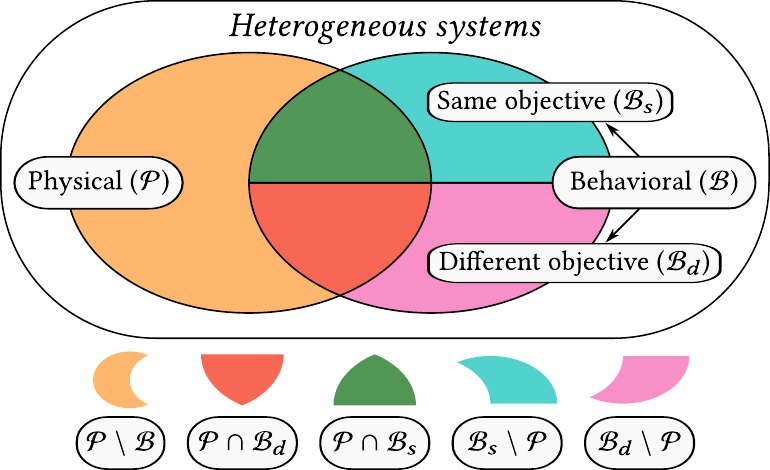}
    \caption{Taxonomy of heterogeneous multi-robot/agent systems. \textit{Top}: the three heterogeneity classes ($\mathcal{P}$, $\mathcal{B}_s$, $\mathcal{B}_d$). \textit{Bottom}: the five mutually exclusive heterogeneity subclasses. Every heterogeneous system belongs to one of these subclasses.}
    \label{fig:taxomony_het}
\end{figure}

% State of the art
Most cooperative MARL works constrain agents to share policy neural network parameters to improve training sample efficiency~\cite{gupta2017cooperative,rashid2018qmix,sukhbaatar2016learning}. This causes the agents' models to be identical and, thus, homogeneous. While this is beneficial to speed-up training, it can prevent learning in scenarios that require heterogeneous behavior. A classical method of overcoming this imposed homogeneity is to include a unique integer (e.g., the agent's index) as part of each agent's observations~\cite{foerster2016learning,gupta2017cooperative}. This allows the agents to share the same policy while exhibiting \textit{apparently} different behavior. Despite its wide adoption, this solution has many drawbacks~\cite{christianos2021scaling}. 

We are interested in learning \textit{truly} heterogeneous decentralized MARL policies. While it is common practice to learn heterogeneous policies when optimizing for different objectives~\cite{lowe2017multi}, there is a dearth of work in applying this paradigm to scenarios where the objective is shared. Current solutions are few and tailored to specific tasks, and, as such, do not address the broader study and categorization of heterogeneity in MARL. Furthermore, they are limited to noise-free videogame-like MARL benchmarks~\cite{samvelyan2019starcraft,kurach2020google}, without considering real-world multi-robot tasks with inter-agent communication.
Therefore, we need a framework that enables \textit{true} heterogeneity among communicating MARL agents and can learn policies that run in a decentralized fashion for (real-world) heterogeneous multi-robot systems. 

% Contributions
In this work, we introduce Heterogeneous Graph Neural Network Proximal Policy Optimization (HetGPPO), a paradigm for heterogeneous MARL that overcomes the aforementioned issues. 
HetGPPO is a framework for training heterogeneous MARL policies that leverages a Graph Neural Network (GNN) for differentiable inter-agent communication.
Our architecture enables learning for heterogeneous agents while being conditioned only on local communication and local observations. This enables to train HetGPPO in a decentralized fashion, in-line with the Decentralized Training Decentralized Execution (DTDE) paradigm~\cite{jaques2019social}. 

We begin by presenting a taxonomy of heterogeneous systems in \autoref{sec:taxonomy}. The purpose of this taxonomy is to classify such systems according to the source of their heterogeneity. We use this taxonomy in \autoref{sec:related} to categorize related works in the domains of multi-robot systems and MARL. \autoref{sec:formulation} formulates the MARL problem tackled in this paper. In \autoref{sec:model}, we introduce HetGPPO and its homogeneous counterpart GPPO.
To motivate the need for policy heterogeneity, we distill and define the techniques that homogeneous models use to emulate heterogeneous behavior (\autoref{sec:typing}). Through example scenarios, we demonstrate how these technique work and how they can prove brittle in real-world conditions when compared to truly heterogeneous models. Finally, in \autoref{sec:evaluations}, we present evaluations of our framework both in simulated and real-world multi-robot cooperative scenarios. These show that: (i) when homogeneous methods fail due to strong heterogeneous requirements, HetGPPO succeeds, and, (ii) when homogeneous methods are able to learn apparently heterogeneous behaviors, HetGPPO achieves higher resilience to both training and deployment noise. Furthermore, our real-robot experiments demonstrate how heterogeneous policies are intrinsically more resilient to real-world conditions.
%To the best of our knowledge, this work is the first to not only demonstrate but also exploit the power of heterogeneity in MARL.

In this paper, we demonstrate the power of heterogeneous MARL applied to real-world multi-robot systems.
%\aj{To the best of our knowledge, this work is the first to take a systematic approach towards modeling heterogeneity.} 
%\fixit{on 2nd thoughts: no, i don't think so. balch as also systematic about it.}
We claim the following key contributions:
\begin{enumerate}
\item A taxonomy of heterogeneous systems that jointly categorizes research in the multi-robot and multi-agent domains;
\item A discourse on \textit{behavioral typing} techniques that homogeneous models rely on to emulate heterogeneous behavior, with empirical evidence for their brittleness in deployment;
\item HetGPPO, a MARL model able to learn heterogeneous communicating policies in a decentralized fashion; and,
\item Detailed evaluations of the performance and resilience of heterogeneous policies compared to homogeneous ones in several cooperative multi-robot tasks, both through simulations and real-world experiments.
\end{enumerate}

% \begin{enumerate}
% \item A taxonomy of heterogeneous systems \sout{with the goal of providing} that provides a shared framework to categorize heterogeneous systems research;
% \item \aj{A discourse on} \textit{behavioral typing} techniques that homogeneous models rely on to emulate heterogeneous behavior, with empirical evidence for their brittleness in simulation and in the real world;
% \item HetGPPO, a MARL model able to learn heterogeneous communicating policies in a decentralized fashion; and,
% \item Detailed evaluations of the performance and resilience of heterogeneous policies compared to homogeneous ones through \sout{evaluations in} several cooperative multi-robot tasks, both in simulations and through real-world experiments.
% \end{enumerate}

%%%%%%%%%%%%%%%%%%%%%%%%%%%%%%%%%%%%%%%%%%%%%%%%%%%%%%%%%%%%%%%%%%%%%%%%%%%%%%%%
\section{Taxonomy of heterogeneous systems}
\label{sec:taxonomy}

In spite of a substantial body of work attempting to stimulate research on heterogeneous systems (see~\cite{ayanian2019dart} and the references therein), %and work and to create heterogeneity measures~\cite{balch2000hierarchic, twu2014measure, hu2022policy}, 
the robotics and learning community still lacks a shared and structured taxonomy of heterogeneous systems. To properly characterize the related works in the heterogeneity (diversity)
%\footnote{In this work, we use the terms `heterogeneity' and `diversity' interchangeably.}
domain, we introduce a taxonomy of heterogeneous systems, shown in \autoref{fig:taxomony_het}. According to our taxonomy, system heterogeneity is categorized in two classes: \textit{Physical} ($\mathcal{P}$) and \textit{Behavioral} ($\mathcal{B}$).

A team is considered \textit{physically} ($\mathcal{P}$) heterogeneous when at least one of its components (i.e., agents, robots) differs from the others in terms of hardware or physical constraints. That is, it might have different sensors, actuators, motion constraints, etc. These physical differences might lead to different capabilities. For example, a small drone might be able to fly and move aggressively, but likely has shorter battery life than a big and slow ground robot. This type of heterogeneity can lead to different observation and action spaces in the context of learning, for example when robots are equipped with different sensors or actuators.

A team is considered \textit{behaviorally} ($\mathcal{B}$) heterogeneous when at least one of its components differs from the others in terms of software or behavioral model. That is, two behaviorally heterogeneous agents can produce distinct policy outputs when observing the same input. 
For example, two physically identical drones might cooperate to monitor a site: here, one drone can survey from far away and direct the other to areas that need closer inspection. Behavioral heterogeneity is divided in two: \textit{Same objective} ($\mathcal{B}_s$) and \textit{Different objective} ($\mathcal{B}_d$). In $\mathcal{B}_s$ heterogeneous systems, agents optimize the same objective function through heterogeneous behavior. 
In MARL, this means that they share the same (global or local) reward function. $\mathcal{B}_s$ heterogeneous systems usually represent cooperative settings~\cite{chenghao2021celebrating}. However, they could also model adversarial scenarios where agents with the same objective compete for limited resources~\cite{blumenkamp2021emergence}. 
In $\mathcal{B}_d$ heterogeneous systems, agents optimize different objective functions through heterogeneous behavior. In MARL, this means that they have different local reward functions or a global reward deriving from the composition of such local functions. $\mathcal{B}_d$ heterogeneous systems usually represent non-cooperative or adversarial settings~\cite{lowe2017multi}. However, they could also model cooperative scenarios where agents optimize different sub-functions for a higher-order task~\cite{christianos2021scaling}. For example, in cooperative search and rescue scenarios, one robot might only be tasked to remove debris, while the others are tasked with the search in an uncluttered space.

Physical and behavioral heterogeneity are not mutually exclusive. Thus, the three heterogeneity classes introduced ($\mathcal{P}$, $\mathcal{B}_s$, $\mathcal{B}_d$) delineate five heterogeneity subclasses that a system can belong to: 

\begin{itemize}
    \item $\mathcal{P} \setminus \mathcal{B}$: Agents are physically different but share the same behavioral model.
    \item $\mathcal{P} \cap \mathcal{B}_d$: Agents are physically different and differ in behavioral models and objectives.
    \item $\mathcal{P} \cap \mathcal{B}_s$: Agents are physically different and differ in behavioral models, but share the same objective.
    \item $\mathcal{B}_s \setminus \mathcal{P}$: Agents are physically identical and share the same objective but differ in behavioral models.
    \item $\mathcal{B}_d \setminus \mathcal{P}$: Agents are physically identical but differ in behavioral models and objectives.
\end{itemize}

While this taxonomy is concerned with classifying heterogeneous systems, it does not attempt to measure the \textit{degree} of heterogeneity. Furthermore, it represents a \textit{high-level} classification and does not consider dynamic $\mathcal{P}$ heterogeneity, such as different battery levels or hardware deterioration~\cite{shang2014swarm}.

% also we know that p heterogeneity can be dynamic (i.e. ) 
%\fixit{Not sure we need the following. TBD.}
%Heterogeneity classes can be furtherly divided into more refined subclasses to create a more detailed version of this taxonomy. Due to the scope of this paper, we leave this to future work.

%%%%%%%%%%%%%%%%%%%%%%%%%%%%%%%%%%%%%%%%%%%%%%%%%%%%%%%%%%%%%%%%%%%%%%%%%%%%%%%%
\section{RELATED WORK}
\label{sec:related}

\begin{table}[tb]
\caption{Related work in heterogeneous multi-robot/agent systems classified according to our taxonomy of \autoref{sec:taxonomy}.}
\label{tab:related_works_taxonomy} 
\resizebox{\linewidth}{!}{%
\begin{tabular}{r >{\centering}p{3cm} >{\centering\arraybackslash}p{3cm}}
\toprule
 Heterogeneity class & \rotatebox{0}{Multi-robot systems} & \rotatebox{0}{MARL} \\
\midrule
$\mathcal{P} \setminus\mathcal{B}$ & \parbox{3cm}{\centering\cite{boroson20193d}} &\parbox{3cm}{\centering\cite{wakilpoor2020heterogeneous},\cite{terry2020revisiting}} \\
\midrule
$\mathcal{P} \cap \mathcal{B}_d$ & \parbox{3cm}{\centering\cite{michael2014collaborative}} &\parbox{3cm}{\centering\cite{lowe2017multi},\cite{christianos2021scaling}} \\
\midrule
$\mathcal{P} \cap \mathcal{B}_s$ & 
\parbox{3cm}{\centering\cite{emam2020adaptive},\cite{notomista2019optimal},\cite{emam2021data},\cite{mayya2021resilient},\cite{notomista2021resilient},\\\cite{prorok2017impact},\cite{pimenta2008sensing},\cite{santos2018coverage},\cite{kim2022coverage},\cite{malencia2022adaptive},\\\cite{manjanna2018heterogeneous},\cite{chand2013mapping},\cite{debord2018trajectory}}
 & \parbox{3cm}{\centering\cite{seraj2022learning}}\\
\midrule
$\mathcal{B}_s \setminus \mathcal{P}$  & \parbox{3cm}{\centering\cite{ayanian2019dart},\cite{berman2007bio},\cite{balch2000hierarchic},\cite{balch1997learning},\cite{li2004learning},\\\cite{spica2020real},\cite{wang2021game}} & \parbox{3cm}{\centering\cite{wang2020roma},\cite{chenghao2021celebrating},\cite{wang2021rode}}\\
\midrule
$\mathcal{B}_d \setminus \mathcal{P}$ &  \parbox{3cm}{\centering\cite{goldberg1997interference},\cite{schneider1998territorial}} & \parbox{3cm}{\centering\cite{lowe2017multi},\cite{christianos2021scaling}}\\
\bottomrule
\end{tabular}}
\end{table}

In this section, we review the current state of the art in the area of heterogeneous multi-robot/agent systems. %\matteo{Works from the heterogeneous multi-robot systems domain are discussed in~\autoref{sec:het_in_robotics}, while works from the MARL domain are discussed in~\autoref{sec:het_in_marl}.}
We classify the related works according to our taxonomy in \autoref{tab:related_works_taxonomy}.

\subsection{Heterogeneity in multi-robot systems}
\label{sec:het_in_robotics}

The core literature on heterogeneous robotics has generally focused on developing coordination algorithms that leverage the physical heterogeneity of a team to their advantage.
Therefore, these works fall in the  $\mathcal{P} \cap \mathcal{B}_s$ class.
Such diversity can manifest itself in the form of different sensor ranges~\cite{pimenta2008sensing}, diverse sensing capabilities~\cite{santos2018coverage}, or different maximum speeds~\cite{kim2022coverage}.
These differences can then be exploited in a variety of problems such as multi-robot coverage~\cite{pimenta2008sensing,santos2018coverage,kim2022coverage} and heterogeneous task assignment~\cite{notomista2019optimal, prorok2017impact}, with resilient formulations that can handle uncertainties in robot capabilities~\cite{emam2020adaptive} or the environment~\cite{notomista2021resilient, mayya2021resilient,emam2021data}.
Sensor heterogeneity has also received attention in the context of active sampling and mapping~\cite{manjanna2018heterogeneous,malencia2022adaptive}, where heterogeneous computational resources can impact task execution~\cite{chand2013mapping}.
Lastly,  $\mathcal{P} \cap \mathcal{B}_s$ diversity has also been investigated in more complex problems such as heterogeneous trajectory planning~\cite{debord2018trajectory}. 

Interestingly, such physical diversity without behavioral diversity ($\mathcal{P} \setminus \mathcal{B}$) can often represent a \textit{constraint} for the problem. Works in this heterogeneity class
try to behaviorally reconcile the physical heterogeneity of robots in order to apply homogeneous solutions to the problem at hand.  Heterogeneous multi-robot SLAM is an example application where scans coming from different robots, equipped with diverse sensors, need to be matched in order to build a homogeneous shared map~\cite{boroson20193d}.

% Heterogeneous task assignment is concerned with assigning a team of robots to different tasks according to their capabilities~\cite{notomista2019optimal}. Different resilient formulations have been proposed to deal with uncertainty in robot capabilities~\cite{emam2020adaptive} and in the environment~\cite{notomista2021resilient, mayya2021resilient,emam2021data}. Heterogeneous assignment can also be tackled from a macroscopic swarm perspective in a decentralized fashion~\cite{prorok2017impact}. Heterogeneous multi-robot coverage is another traditional problem where robots have to distribute in an environment according to a given density function. In this domain, physical heterogeneity can derive from different sensor ranges~\cite{pimenta2008sensing}, diverse sensing capabilities~
% \cite{santos2018coverage}, or different maximum speeds~\cite{kim2022coverage}. Sensor heterogeneity has been considered also in the context of active sampling and mapping~\cite{manjanna2018heterogeneous,malencia2022adaptive}, where heterogeneous computational resources can impact task execution~\cite{chand2013mapping}. Lastly,  $\mathcal{P} \cap \mathcal{B}_s$ diversity has also been investigated in more complex problems such as heterogeneous trajectory planning~\cite{debord2018trajectory}.

Behavioral heterogeneity for physically identical robots is a less explored but promising research direction~\cite{ayanian2019dart}.
Works in this area mostly tackle cooperative problems, leveraging  $\mathcal{B}_s \setminus \mathcal{P}$ heterogeneity.
%Early research results from studying behavioral roles in ant colony systems~\cite{dorigo2006ant,berman2007bio}.
Early research by Balch~\cite{balch1997learning,balch2000hierarchic} and Li et al.~\cite{li2004learning} focuses on learning behavioral specialization for multi-robot teams using RL. 
Game-theoretic autonomous racing~\cite{spica2020real, wang2021game} constitutes an adversarial setting of $\mathcal{B}_s \setminus \mathcal{P}$ heterogeneity. Note that game-theoretic controllers do not present heterogeneous behavior when all players use the symmetric Nash equilibrium strategy~\cite{nash1950equilibrium}. However, heterogeneity emerges when some robots in the team  use traditional model predictive controllers.

%\matteo{Heterogeneous behavior with different objectives ($\mathcal{B}_d$) has been analyzed in cooperative robotics. In~\cite{goldberg1997interference, schneider1998territorial}, the authors show that a global task can be divided into several sub-tasks and assigned to different sub-groups of identical robots, resulting in $\mathcal{B}_d \setminus \mathcal{P}$ heterogeneity. When the robots \amanda{additionally?} have physical differences between sub-groups, these differences can be leveraged to tackle complex multi-robot tasks. In~\cite{michael2014collaborative}, the authors deploy $\mathcal{P}$ heterogeneous robots with different objectives to collaboratively map an earthquake-damaged building, resulting in  $\mathcal{P} \cap \mathcal{B}_d$ heterogeneity.}

Conversely, heterogeneous behavior with different objectives ($\mathcal{B}_d$) has also been analyzed for cooperative robotic tasks, for instance, by dividing a global task into sub-tasks for groups of identical robots ($\mathcal{B}_d \setminus \mathcal{P}$)~\cite{goldberg1997interference, schneider1998territorial}.
When the robots additionally have physical differences between sub-groups, these differences can be leveraged to tackle complex multi-robot tasks, such as post-disaster collaborative mapping~\cite{michael2014collaborative}, resulting in  $\mathcal{P} \cap \mathcal{B}_d$ heterogeneity.

All the works discussed in this subsection focus on a given heterogeneity class and problem, and develop a targeted solution for that setting.
To a large extent, the approaches leverage conventional control theoretical methods. 
Our work, in contrast, proposes a learning-based framework to synthesize communicating multi-agent/robot policies, and can be applied to any heterogeneity class.

% Heterogenous behaviour:
% nora dart~\cite{ayanian2019dart}
% bio-inspired~\cite{dorigo2006ant,berman2007bio}
%balch~\cite{balch1997learning,balch2000hierarchic}
%goldberg~\cite{goldberg1997interference, schneider1998territorial}
%martinoli stick pulling~\cite{li2004learning}

%Heterogenous SLAM~\cite{boroson20193d}
%Heterogenous trajectory planning~\cite{debord2018trajectory}

%Heterogenous mapping and sampling:
%Heterogenous sensisg for active sampling~\cite{malencia2022adaptive}
%heterogenous computational for mapping~\cite{chand2013mapping}
%heterogenous sensing for monitoring and sampling~\cite{manjanna2018heterogeneous}

%Heterogeneous coverage:
%Coverage with different sensor ranges~\cite{pimenta2008sensing}
%heterogeneous sensing capabilities (density functions)~\cite{santos2018coverage}
%different speeds~\cite{kim2022coverage}

%Heterogeneous assignment:
%prorok continuous assignment~\cite{prorok2017impact}
%data-driven heterogeneous robust~\cite{emam2021data}
%energy-based with known capabilities~\cite{notomista2019optimal}
%energy-based with unknown and dynamic %capabilities~\cite{emam2020adaptive}
%resilient and energy-based with given %capabilities~\cite{notomista2021resilient}
%resilient with unknown environment~\cite{mayya2021resilient}

\subsection{Heterogeneity in MARL}
\label{sec:het_in_marl}

MARL has recently gained increasing traction as an effective technique to tackle multi-robot problems~\cite{zhang2021multi}. Using MARL, it is possible to synthesize efficient decentralized multi-agent controllers for hard coordination problems~\cite{bernstein2002complexity}.
Homogeneous policies (that share parameters) for physically identical agents are abundant in MARL~\cite{gupta2017cooperative,rashid2018qmix,foerster2018counterfactual,kortvelesy2022qgnn,sukhbaatar2016learning} and constitute the core of the research literature.
%\matteo{Parameter sharing allows agents to benefit from collective experiences and thereby reduces training time.
%On the other hand, it enforces centralized training and constraints agents' policies to be identical (i.e., homogeneous). }
In an attempt to emulate heterogeneous behavior, a common practice is to augment each agent's observation space with a unique index that represents the agent's type~\cite{foerster2016learning,gupta2017cooperative}.
In this case, agents share the same homogeneous multimodal policy, conditioned on a unique constant index.
We define and discuss in depth the limitations of this approach in \autoref{sec:typing}.
$\mathcal{P} \setminus \mathcal{B}$ heterogeneity in MARL focuses on leveraging the power of parameter sharing and homogeneous training for physically different agents. This is achieved by mapping heterogeneous observation spaces into homogeneous fixed-length encodings~\cite{wakilpoor2020heterogeneous}, or by padding and including the agent index into observations~\cite{terry2020revisiting}.

The majority of heterogeneous MARL literature falls in the $\mathcal{B}$ heterogeneity class.
Different behavioral roles for physically identical agents can be learned through various techniques, such as conditioning agents' policies on a latent representation~\cite{wang2020roma}, decomposing and clustering action spaces~\cite{wang2021rode}, or by an intrinsic reward that maximizes the mutual information between the agent's trajectory and its role~\cite{chenghao2021celebrating}.
%#\matteo{In ROMA~\cite{wang2020roma}, emergent roles for identical agents are learnt during the training phase. Roles are modeled through a latent representation and agents' policies are conditioned upon them. Similarly, in \cite{chenghao2021celebrating}, the mutual information between an agent's trajectory and its role is maximized during training. This is done through the introduction of an intrinsic reward.
%Furthermore, each agent keeps a local non-shared critic and a shared one.
%RODE~\cite{wang2021rode} constructs roles by decomposing the action spaces and clustering actions according to their effects on the environment.}
All the aforementioned works consider physically identical agents with the same objective, thus leveraging $\mathcal{B}_s \setminus \mathcal{P}$ heterogeneity.
Furthermore, they do not use inter-agent communication, and hence their application to highly partially observable coordination problems is limited.
When considering physically different robots, heterogeneous action or observation spaces have to be taken into account.
Such $\mathcal{P} \cap \mathcal{B}_s$ heterogeneity with communicating agents can be modeled, for instance, by an ad-hoc GNN layer for each physically different robot type~\cite{seraj2022learning}.
While this may be suitable for some tasks where robot types are known beforehand, it prevents physically identical agents from learning heterogeneous behavior.

Behavioral heterogeneity with different objectives ($\mathcal{B}_d$) emerges due to different agent reward functions, as discussed in \autoref{sec:taxonomy}. MADDPG~\cite{lowe2017multi} uses this paradigm in a centralized training approach to learn individual (not shared) actors and critics. They test their approach in mixed cooperative-competitive tasks. In these tasks, both physically identical and physically different agents (i.e., different maximum speeds) are considered.  
Thus, MADDPG leverages heterogeneity classes  $\mathcal{B}_d \setminus \mathcal{P}$ and $\mathcal{P} \cap \mathcal{B}_d$. The same heterogeneity classes are studied in~\cite{christianos2021scaling}, which proposes to use parameter sharing among sub-groups of agents which are physically identical and share the same reward function. This approach, however, prevents physically identical agents with the same objective to employ different behavioral roles to solve a task.

Most works discussed in this section propose solutions to problems that sit exclusively within one given heterogeneity subclass. 
While a selected few could be applied to multiple classes~\cite{lowe2017multi,wang2020roma,chenghao2021celebrating}, they leverage centralized training methods and do not consider inter-agent communication. These are two key features needed to make MARL suitable for multi-robot problems.

\section{Problem formulation}
\label{sec:formulation}

% put robot
We now formulate the multi-robot MARL problem tackled in this work. To do so, we first introduce the multi-agent extension of a Partially Observable Markov Decision Process (POMDP)~\cite{kaelbling1998planning}.

\textbf{Partially Observable Markov Games}. A Partially Observable Markov Game (POMG) is defined as a tuple
$$\left \langle \mathcal{V}, \mathcal{S}, \mathcal{O}, \left \{ \sigma_i \right \}_{i \in \mathcal{V}},  \mathcal{A}, \left \{ \mathcal{R}_i \right \}_{i \in \mathcal{V}}, \mathcal{T}, \gamma \right \rangle,$$
where $\mathcal{V} = \{1,\ldots, n\}$ denotes the set of agents,
$\mathcal{S}$ is the state space, shared by all agents, and,
$\mathcal{O} \equiv \mathcal{O}_1 \times \ldots \times \mathcal{O}_n$ and
$\mathcal{A} \equiv \mathcal{A}_1 \times \ldots \times \mathcal{A}_n$
are the observation and action spaces, with $\mathcal{O}_i \subseteq \mathcal{S}, \; \forall i \in \mathcal{V}$. 
Further, $\left \{ \sigma_i \right \}_{i \in \mathcal{V}}$ 
and
$\left \{ \mathcal{R}_i \right \}_{i \in \mathcal{V}}$
are the agent observation and reward functions\footnote{Note that, while we formulate our problem with local agent reward functions $\mathcal{R}_i$ (to enable learning of $\mathcal{B}_d$ heterogeneity), our experiments all present a global reward function $\mathcal{R} = \mathcal{R}_1=\ldots=\mathcal{R}_n$ which cannot be decomposed into local sub-functions. This reward encodes a global cooperative objective, leading to $\mathcal{B}_s$ heterogeneity.}, such that
$\sigma_i : \mathcal{S} \mapsto \mathcal{O}_i$, and,
$\mathcal{R}_i: \mathcal{S} \times \mathcal{A} \times \mathcal{S} \mapsto \R$.
$\mathcal{T}$ is the stochastic state transition model, defined as $\mathcal{T} : \mathcal{S} \times \mathcal{A} \times \mathcal{S} \mapsto [0,1]$.
Lastly, $\gamma$ is the discount factor.

We structure the agents in a communication graph $\mathcal{G} = \left (\mathcal{V},\mathcal{E} \right )$. Nodes $i \in \mathcal{V}$ represent agents and edges $e_{ij} \in \mathcal{E}$ represent communication links. The set of edges is dependent of the maximum agent communication range and changes over time. The communication neighborhood of each agent is defined as $\mathcal{N}_i \equiv \{v_j \, | \, e_{ij} \in \mathcal{E}\}$. 

At each timestep $t$, each agent $i$ gets an observation $o_i^t = \sigma_i  (s^t  ) \in \mathcal{O}_i$ that is a portion of the global state $s^t \in \mathcal{S}$.
This is communicated to the neighboring agents $\mathcal{N}_i^t$. A stochastic policy $\pi_i$ uses this information to compute an action $a_i^t \sim \pi_i  (\cdot \rvert o_{\mathcal{N}_i}^t  )$. The agents' actions $\mathbf{a}^t =  (a_1^t,\ldots,a_n^t  ) \in \mathcal{A}$, along with the current state $s^t$, are then used in the transition model to obtain the next state $s^{t+1} \sim \mathcal{T} \left (\cdot \rvert s^t,\mathbf{a}^t \right )$. A reward $r_i^t = \mathcal{R}_i  \left (s^t,\mathbf{a}^t,s^{t+1} \right )$ is then fed to agent $i$. 

The goal of each agent is to maximize the sum of discounted rewards $ v_i^t = \sum_{k=0}^{T} \gamma^k r_{i}^{t+k}  $ over an episode with horizon $T$, potentially infinite\footnote{$\gamma^k$ indicates $\gamma$ to the power of $k$, and not the timestep superscript.}. $v_i^t$ is called the return. Each agent has a value function $V_i(o_{\mathcal{N}_i}) =  \mathbb{E}_{\pi_i}\left [ v_i^t  \big\rvert  o_{\mathcal{N}_i}^t = o_{\mathcal{N}_i}\right ]$, which represents the expected return starting from observations $o_{\mathcal{N}_i}$ and following policy $\pi_i$. This function estimates the ``goodness'' of an observation. In this work, we use the Proximal Policy Optimization (PPO) actor-critic algorithm~\cite{schulman2017proximal}, which approximates the policy (actor) and the value function (critic) using neural networks and a constrained policy gradient update.

% \textbf{Goal}.
%  (i) \amanda{first, we need the method, which allows us to ... ; this is formulated as:} The goal of this work is to learn heterogeneous policies $\pi_i ( o_{\mathcal{N}_i}^t; \theta_i  )$ and critics $V_i(o_{\mathcal{N}_i}; \theta_i)$ conditioned on the neural network parameters $\theta_i$, different for each agent. The observations $o_{\mathcal{N}_i}^t$ are obtained through a differentiable communication channel. 
%  (ii) next, we use this method to verify whether, and ...  

\textbf{Problem}.
Learn heterogeneous policies $\pi_i ( o_{\mathcal{N}_i}^t; \theta_i )$ and critics $V_i(o_{\mathcal{N}_i}; \theta_i)$ conditioned on the neural network parameters $\theta_i$, different for each agent.
The observations $o_{\mathcal{N}_i}^t$ from the agent's neighborhood $\mathcal{N}_i$ are obtained through a differentiable communication channel, making learning inherently decentralizable.

Our objective is to crystallize the role of heterogeneity in MARL policies. Towards this end, we develop a model that addresses the problem description above, motivating it with an empirically-backed discourse on the shortcomings of homogeneous policies (and the \textit{behavioral typing} techniques that they rely on).
%by defining the \textit{behavioral typing} techniques these employ}
%Using this model, we will then construct a discourse on the shortcomings of homogeneous methods by defining the \textit{behavioral typing} techniques these employ.

%For problem statement, if this is still unresolved: we could say "a model that addresses the problem above, and support our constructs with an empirically.."

%%%%%%%%%%%%%%%%%%%%%%%%%%%%%%%%%%%%%%%%%%%%%%%%%%%%%%%%%%%%%%%%%%%%%%%%%%%%%%%%
\section{Heterogeneous model}
\label{sec:model}

We introduce the two MARL models that constitute the methodology leveraged in this work: Graph Neural Network Proximal Policy Optimization (GPPO) and its heterogeneous counterpart, HetGPPO. 

GPPO builds upon Independent Proximal Policy Optimization (IPPO)~\cite{de2020independent}. In IPPO, each agent learns a local critic $V_i(o_i)$ and actor $\pi_i(o_i)$, conditioned only on its own observations.
%By conditioning the critic only on local observations and not on the full state $s$, non-stationarity is introduced during training.
Conditioning the critic only on local observations and not on the full state $s$ introduces non-stationarity during training.
This results in other agents being considered as part of the environment and not explicitly modeled in the critic. While this can be problematic, it has the advantage of not requiring global information during training. Furthermore, IPPO has been shown to outperform many fully-observable critic models on state-of-the-art MARL benchmarks~\cite{de2020independent}. 

GPPO overcomes the limitations of IPPO while maintaining its benefits. It uses a GNN communication layer, allowing agents to share information in neighborhoods to coordinate and overcome partial observability. Thanks to this, the GPPO critic $V_i(o_{\mathcal{N}_i})$ and actor $\pi_i(o_{\mathcal{N}_i})$ are conditioned on local communication neighborhood observations $o_{\mathcal{N}_i}$. This helps overcome non-stationarity, while requiring only local information and communication during training. 

\begin{figure}[tb]
\centering
\includegraphics[width=\linewidth]{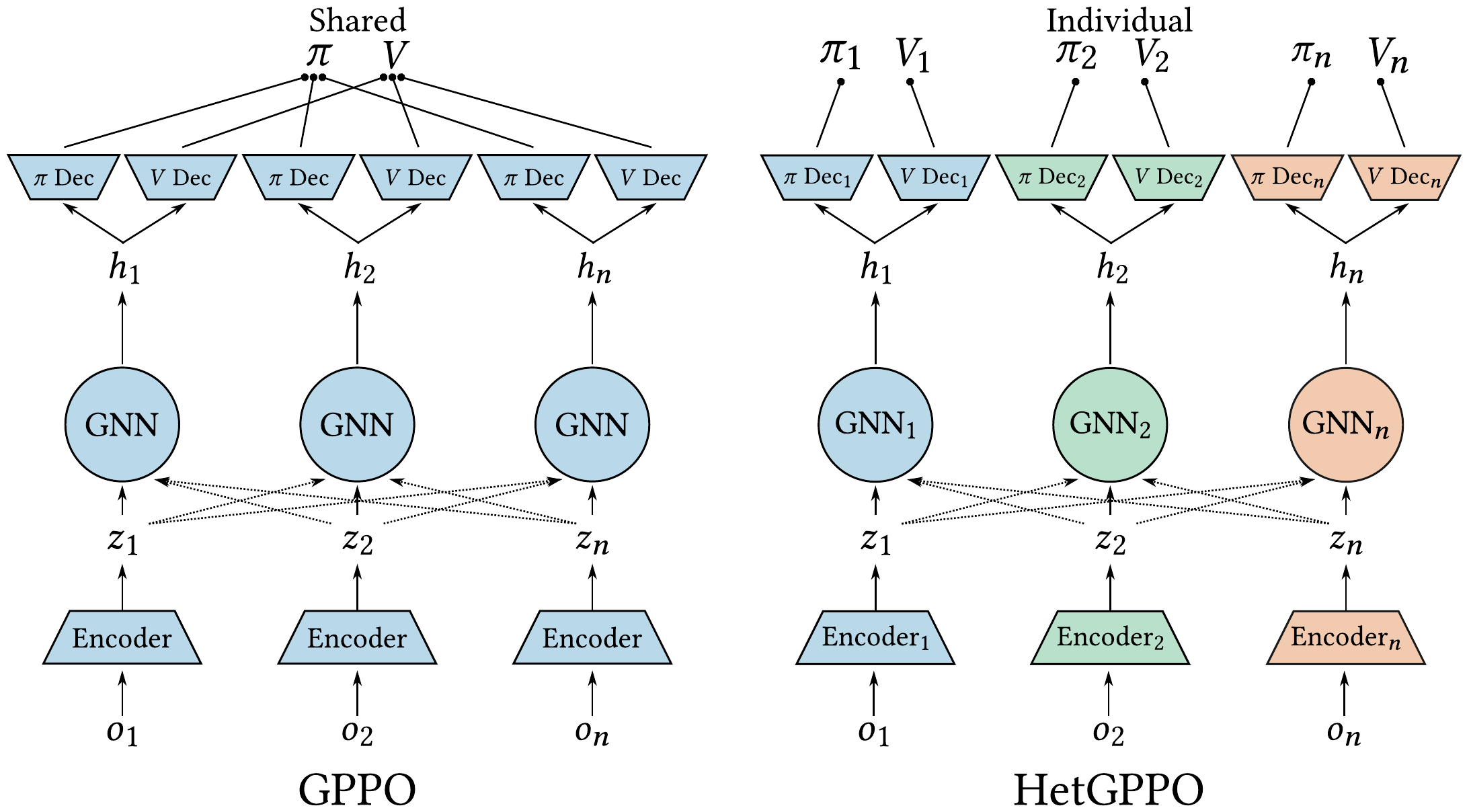}
\caption{Architecture of GPPO and HetGPPO: MARL models with communicating agents. Each agent passes its observation through an encoder, then aggregates messages received from its neighbors using a translation-invariant message-passing GNN and updates its hidden state $h_i$. $h_i$ is then used as input to the policy and value decoders (Dec). HetGPPO is equivalent to GPPO without parameter sharing.}
\label{fig:gppo}
\end{figure}

The GPPO model is illustrated in \autoref{fig:gppo}. At each timestep, each agent $i$ observes the environment, collecting the observations $o_i$. These observations contain absolute geometrical features, such as the agent position $\mathbf{p}_i \in \R^2$. The non-absolute features are passed through a Multi Layer Perceptron (MLP) encoder, obtaining an embedding $z_i$. The absolute position and the agent velocity $\mathbf{v}_i \in \R^2$ are used to compute edge features $e_{ij}$, which are relative features of agents $i$ and $j$. In this work, we use the relative position $\mathbf{p}_{ij} = \mathbf{p}_i - \mathbf{p}_j$ and relative velocity $\mathbf{v}_{ij} = \mathbf{v}_i - \mathbf{v}_j$ as edge features $e_{ij} = \mathbf{p}_{ij} \left |\right | \mathbf{v}_{ij}$, where $\left |\right |$ indicates the concatenation operation. This process ensures that GNN outputs are invariant to translations in $\R^2$ (i.e., the same output is obtained if all the team is translated in space), helping the model generalize~\cite{fuchs2020se}. The edge features $e_{ij}$ and the agent embedding $z_i$ are then used in the message-passing GNN kernel:

$$
h_i = \psi_{\scriptscriptstyle \theta_i}(z_i) + \bigoplus_{j \in \mathcal{N}_i}\phi_{\scriptscriptstyle \theta_i}(z_j \left |\right | e_{ij}).
$$

Here, $\psi_{\scriptscriptstyle \theta_i}$ and $\phi_{\scriptscriptstyle \theta_i}$ are two MLPs, parameterized by the agent parameters $\theta_i$\footnote{With $\theta_i$ we indicate the parameters for all the neural network layers of agent $i$.}, and $\bigoplus$ is an aggregation operator (e.g., sum).
The GNN output $h_i$ is then fed to two different MLP decoders, which output the action $a_i \sim \pi_i(\cdot|o_{\mathcal{N}_i})$ and the value $V_i(o_{\mathcal{N}_i})$. 
Similar to IPPO, GPPO uses parameter sharing to improve sample efficiency. Thus $\theta_1 = \ldots = \theta_n$. 
Parameter sharing allows agents to benefit from collective experiences and thereby reduces training time.
On the other hand, it enforces centralized training and constraints agents' policies to be identical (i.e., homogeneous). 

HetGPPO, \textit{removes} the parameter sharing constraint of GPPO, thus allowing agent policies to diverge, $\theta_1 \neq \ldots \neq \theta_n$.
However, the impact of not sharing parameters in the context of GNN communications is profound: the permutation equivariance property of GNNs~\cite{xu2018powerful} does not hold, since the agents now learn different message encoding and interpreting strategies.
This results in the GNN having to learn a different team output for all the possible permutations of a given team input, instead of learning only one output. This can lead to decreases in generalization power and sample efficiency.
On the other hand, gradients are backpropagated through communication neighborhoods, enabling agents to learn collectively from local interactions. 

The structure of HetGPPO, shown in \autoref{fig:gppo}, allows for \textit{Decentralized Training Decentralized Execution (DTDE)}.
This is thanks to the fact that GPPO critics are not conditioned on global information.
While GPPO uses parameter sharing, HetGPPO removes this need, thus enabling training in any environment where just inter-agent communication is possible. We note that, by implementing an ad-hoc mechanism to achieve decentralized parameter sharing (e.g., through distributed optimization~\cite{yu2022dinno}), GPPO could be trained in a decentralized fashion as well. 

We implement HetGPPO and GPPO in PyTorch~\cite{paszke2019pytorch} and employ the RLlib~\cite{liang2018rllib} framework for training. The code is available \href{https://github.com/proroklab/HetGPPO}{here}\footnote{\url{https://github.com/proroklab/HetGPPO}}. Simulations are executed in custom created scenarios using the VMAS simulator~\cite{bettini2022vmas}, available at this \href{https://github.com/proroklab/VectorizedMultiAgentSimulator}{link}\footnote{\url{https://github.com/proroklab/VectorizedMultiAgentSimulator}}.

%%%%%%%%%%%%%%%%%%%%%%%%%%%%%%%%%%%%%%%%%%%%%%%%%%%%%%%%%%%%%%%%%%%%%%%%%%%%%%%%$
\section{Behavioral typing}
\label{sec:typing}

%In this section we discuss how homogeneous MARL models can learn heterogeneous behavior and why this can lead to brittle and unreliable policies. In \autoref{sec:indexing}, we define the types of indexing that homogeneous models can leverage to differentiate among agents. Subsequently, in \autoref{sec:case_study}, we present two case studies which illustrate how such indexing works and elucidate its limitations.

HetGPPO, introduced above, allows us to learn \textit{truly} heterogeneous policies. Counter-intuitively, it is also possible to learn \textit{apparently} heterogeneous behavior with homogeneous models like GPPO.
This allows agents to emulate heterogeneous behavior while leveraging the sample efficiency benefits of parameter sharing. A shared model can encompass different behavioral types which are activated by particular combinations of the input observations.
%These input combinations help the model identify which agent type an input has to be assigned to.
For example, if two robots are transporting a package towards a destination, the model can identify if an agent is in the back (further from the goal) and assign it a different behavioral type from that of the agent in the front.
The input observations provide the conditions for the model to assign behavioral types to the agents. 

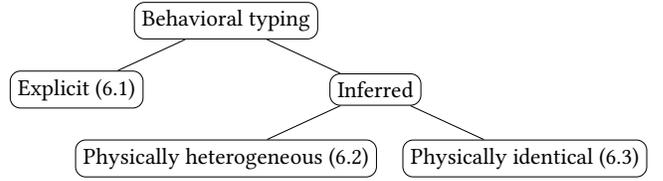
\begin{figure}[tb]
    \centering
    \resizebox{\linewidth}{!}{%
        \begin{tikzpicture}[sibling distance=13em,level distance=3em,
            every node/.style = {shape=rectangle, rounded corners,
            draw, align=center,
            top color=white, bottom color=white!20}]]
            \node {Behavioral typing}
            child { node {Explicit (\ref{sec:explicit_typing})} }
            child { node {Inferred}
              child { node {Physically heterogeneous (\ref{sec:inferred_typing_physical})} }
              child { node {Physically identical (\ref{sec:inferred_typing_identical})} }};
        \end{tikzpicture}}
    \caption{
    Different forms of behavioral typing. Homogeneous policies use typing to differentiate among agents and emulate heterogeneous behavior.}
    \label{fig:homogenous_indexing}
\end{figure}

We refer to this identification process as \textit{typing}.
\autoref{fig:homogenous_indexing} depicts a classification of behavioral typing techniques which we describe in the following subsections.
Note that behavioral types lie in a continuous behavioral space (and are not part of a discrete set)~\cite{balch2000hierarchic}.
%In the following, we define the different ways in which homogeneous models perform behavioral typing. We present a classification of behavioral typing strategies in \autoref{fig:homogenous_indexing}.

\subsection{Explicit behavioral typing}
\label{sec:explicit_typing}
The most popular form of behavioral typing consists in feeding the index $i$ of the agent explicitly as part of the observation. This practice has been used extensively in the MARL literature~\cite{foerster2016learning,gupta2017cooperative,christianos2021scaling,terry2020revisiting}. However, it requires the model to learn a multimodal policy, which switches modes based on this integer index. This can lead to discontinuities in the
agents’ policy and has been shown to perform sub-optimally \cite{christianos2021scaling}.

\begin{definition}[Explicit behavioral typing]
Explicit behavioral typing occurs when a shared decentralized MARL policy is able to type agents based on a constant value concatenated to the input, different for each agent.
\end{definition}

When no explicit index is available, a shared policy may still be able to emulate heterogeneous behavior~\cite{deka2021natural}. We refer to this phenomenon as \textit{inferred} behavioral typing. Inferred typing can occur for both physically heterogeneous and {physically} identical agents.

\subsection{Inferred behavioral typing for physically heterogeneous agents}
\label{sec:inferred_typing_physical}

%We now define the concept of inferred behavioral typing for physically heterogeneous agents and present a case study to elucidate its meaning.
We first present a case study of inferred behavioral typing for agents that are physically heterogeneous.

\begin{definition}[Inferred behavioral typing for physically heterogeneous agents]
Inferred behavioral typing for physically heterogeneous agents occurs when a shared decentralized MARL policy is able to type $\mathcal{P}$ heterogeneous agents through their observations.
\end{definition}

\begin{figure*}[tb]
    \centering
    \includegraphics[width=\linewidth]{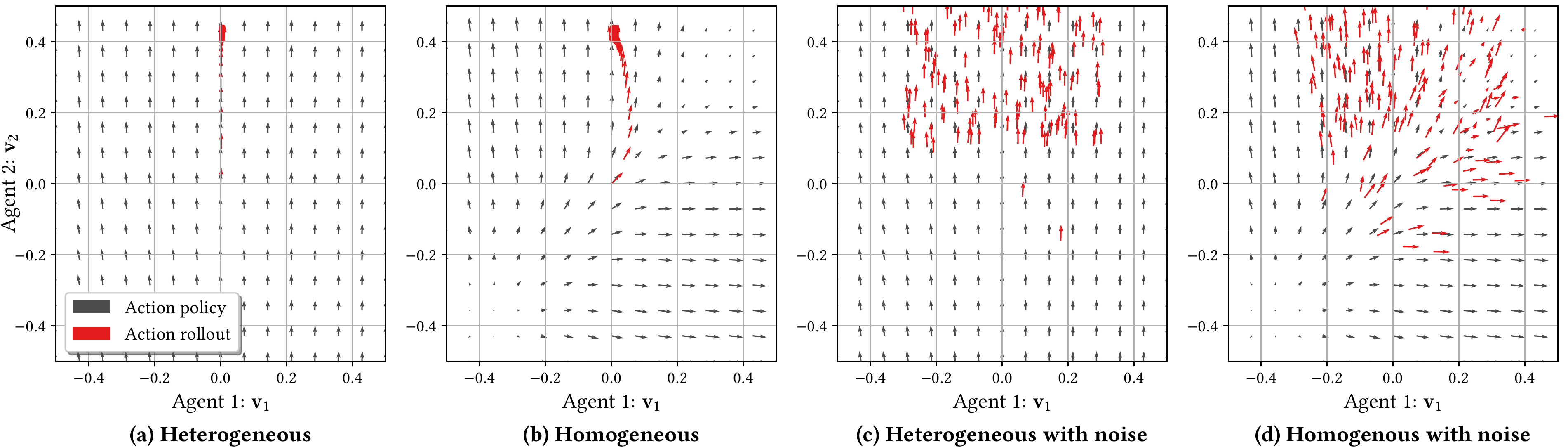}
    \caption{Policies learned for Scenario A represented as vector fields (gray) and rollouts in the environment (red). (a) and (b) are not subject to deployment noise. (c) and (d) are subject to $\pm0.3$ uniform noise on the observations. 
    In these plots, an arrow represents the team action vector $\vec{\mathbf{f}}(\vec{\mathbf{v}}) = [\mathbf{f}_1(\vec{\mathbf{v}}), \mathbf{f}_2(\vec{\mathbf{v}})]$ as a function of the observation $ \vec{\mathbf{v}} = [\mathbf{v}_1, \mathbf{v}_2]$. The rollouts always start in the origin ($\vec{\textbf{v}} = [0,0]$). We can observe how the vector field representing the homogeneous policies is forced to be invariant to permutations of the two inputs and thus is symmetric along $\mathbf{v}_1=\mathbf{v}_2$. This causes it to become brittle in the presence of noise (d), which makes the observations fall in the wrong part of the plane where the symmetry enforces a suboptimal policy (horizontal arrows).}
    \label{fig:vector_field}
\end{figure*}

\textbf{Scenario A (\autoref{fig:vector_field})}.
Consider two robots with different masses, $m_1 > m_2$, located in a 1D workspace at random positions. The robots observe their own position $\mathbf{p}_i \in \R$ and velocity $\mathbf{v}_i \in \R$ and share them via communication. Their action is a force $\mathbf{f}_i \in \R$. They are rewarded collectively to maximize the maximum speed in the team while minimizing the energy consumed.
The optimal policy in this case is, clearly, for the robot with the higher mass to not move at all, while the lighter robot moves at the maximum speed.
Evidently these behaviors are heterogeneous, since $\mathbf{f}_1 \neq \mathbf{f}_2$ when both agents receive the same observations.

% \fixit{Without any observation, only a heterogeneous model could learn them. - might be confusing, since we have observations. Perhaps remove?}

We train the agents in this scenario using GPPO and HetGPPO. 
\autoref{fig:vector_field} shows a graphical representation of the learned policies of each model. 
In these plots, an arrow represents the team action vector $\vec{\mathbf{f}}(\vec{\mathbf{v}}) = [\mathbf{f}_1(\vec{\mathbf{v}}), \mathbf{f}_2(\vec{\mathbf{v}})]$ as a function of the observation $ \vec{\mathbf{v}} = [\mathbf{v}_1, \mathbf{v}_2]$.
A vertical arrow at $[0,0]$ indicates that, when the agents are both still, agent $1$ wants to stay still while agent $2$ wants to increase its velocity.
We plot the policy mean action\footnote{The PPO policy is stochastic and outputs a distribution over $\mathbf{f}_i$.} for every observation pair with a gray vector field, and show a rollout of the policy in red.
In \autoref{fig:vector_field}a, we observe how HetGPPO is able to learn the optimal policy, which is not dependent on any observation.
Thanks to physically inferred typing, GPPO (\autoref{fig:vector_field}b) is surprisingly also able to learn a policy that grants optimal rollouts. We observe how, due to homogeneity, the GPPO policy is forced to be symmetric about the $\mathbf{v}_1 = \mathbf{v}_2$ axis.
Thus, when the GPPO agents are spawned at $[0,0]$, they forcibly take the same action of increasing their velocities.
Due to their $\mathcal{P}$ differences, however, this action produces different velocities, making the rollout quickly diverge from the symmetry and thus producing optimal behavior.
The fact that a physical difference (i.e., different agent mass) produces different observations (i.e., different agent speed) for the same action, enables the homogeneous model to learn an optimal policy with apparent heterogeneous behavior -- an example of \textit{inferred behavioral typing for physically heterogeneous agents}. 

Physically inferred typing proves to be a brittle solution. 
When additive uniform observation noise is injected during rollouts, we observe how the HetGPPO rollout (\autoref{fig:vector_field}c) is not impacted at all, while the GPPO rollout (\autoref{fig:vector_field}d) occasionally falls on the other side of the diagonal, producing the symmetrical opposite of the optimal behavior, and causing the heavy agent to move (horizontal arrows).

\subsection{Inferred behavioral typing for physically identical agents}
\label{sec:inferred_typing_identical}

%We now define the concept of inferred behavioral typing for physically identical agents and present a case study to elucidate its meaning.
We now present a case study of inferred behavioral typing for agents that are physically identical.

\begin{definition}[Inferred behavioral typing for physically identical agents]
Inferred behavioral typing for physically identical agents occurs when a shared decentralized MARL policy is able to type physically identical agents through their observations.
\end{definition}

\begin{figure}[tb]
    \centering
    \begin{subfigure}{0.2\linewidth}
        \hspace*{8pt}
        \includegraphics[height=100pt,keepaspectratio]
        {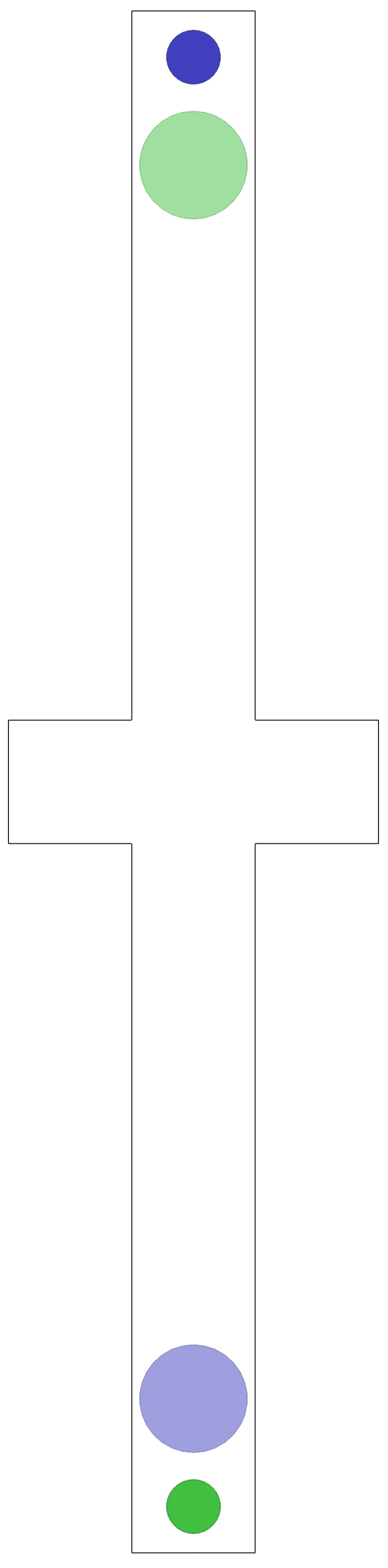}
        \vspace*{13pt}
        \caption{Scenario}
        \label{fig:scenario_b_rendering}
    \end{subfigure}%
    \begin{subfigure}{0.8\linewidth}
        \resizebox{\linewidth}{!}{%
        \input{figures/behavioural_index/give_way.tex}}
          \caption{Training performance}
        \label{fig:scenario_b_training}
    \end{subfigure}%
    \caption{Scenario B. (a): The setup with two robots (bigger circles) on opposite sides of a corridor which need to give way to each other to reach their goals (smaller circles). (b) The training curve for Scenario B, showing that, while the heterogeneous model is able to solve the scenario immediately, homogeneous models need around 300 training iterations to learn inferred behavioral typing for physically identical agents. We plot the mean and standard deviation of 10 different runs. Each iteration is performed over 200 episodes.}
    \label{fig:scenario_b}

\end{figure}

\textbf{Scenario B (\autoref{fig:scenario_b}).}
Consider now two physically identical robots, initialized at different ends of a narrow corridor, depicted in \autoref{fig:scenario_b_rendering}.
Each robot is positioned in front of the other's goal.
The corridor is wide enough to fit only one robot, but contains two robot-sized recesses in the center.
%The scenario is depicted in \autoref{fig:scenario_b_rendering}.
The robots observe and communicate their respective 2D positions and velocities, and are tasked with reaching their goals without colliding.
Thus, the task can only be solved when one robot gives way to the other. 
% This scenario is a slight modification of one available in the VMAS simulator~\cite{bettini2022vmas}. \fixit{careful to not reveal author identity.}

Again, we train the agents with both GPPO and HetGPPO in this scenario. By looking at the training reward plot in \autoref{fig:scenario_b_training}, we can see that both models are able to learn the correct behavior (reward $>700$). GPPO leverages \textit{inferred typing for physically identical agents} and is able to assign the ``give way'' role dynamically according to the relative position and velocity of the two robots.
However, we observe that learning behavioral typing takes all 300 training iterations, while the heterogeneous model learns the optimal solution with only 20 iterations.

%\begin{remark}
%Such role-based indexing allows a homogeneous policy to exhibit distinct behaviors, however, this is reliant on observed cues from the environment and/or the task.
%For instance, an agent needs to observe that it is closer to the recess than the other, and assume the role of side-stepping.
%Such cues may not always be available in a timely manner, or a comparison may not be always possible.
%\end{remark}

%\begin{remark}
%As with physical indexing, behavioral indexing is also extremely prone to environmental noise.
%\aj{Additionally, inferring a behavioural index can be more difficult, since this may be hidden in the structure of the task.}
%In Figure~\ref{fig:concepts-egB}(b), we show the training plots for a homogeneous policy contrasted with a heterogeneous one.
%\fixit{The reward is YY when the task is completed ..}
%We observe that while a heterogeneous policy is quickly able to learn the optimal policy, the homogeneous policy struggles since both agents are attempting to learn the same behavior.
%\hfill $\square$
%\end{remark}

\subsection{Limitations of behavioral typing}

Although homogeneous models can use behavioral typing to learn apparently heterogeneous behavior, this does not prove to be a reliable and scalable solution.

In~\cite{christianos2021scaling} it is shown that the performance of explicit behavioral typing degrades as a function of the number of types to be learned. Furthermore, the authors empirically show that this performance decrease is not related to the capacity of the shared homogeneous model (i.e., the number of parameters).

\begin{figure}[tb]
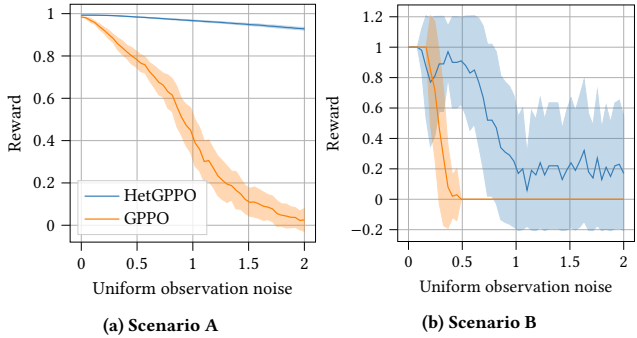

    \centering\hfill
    \begin{subfigure}{0.5\linewidth}
        \resizebox{\linewidth}{!}{%
        \input{figures/physiscal_index/trial.tex}}
        \caption{Scenario A}
        \label{fig:noise_indexing_a}
    \end{subfigure}\hfill
    \begin{subfigure}{0.5\linewidth}
        \resizebox{\linewidth}{!}{%
        \input{figures/behavioural_index/trial.tex}}
        \caption{Scenario B}
        \label{fig:noise_indexing_b}
    \end{subfigure}\hfill
    \caption{Performance of homogeneous and heterogeneous models in the presence of deployment noise on the two inferred typing scenarios. Reward is normalized between 0 and 1. Uniform noise is applied to all observations and it is in the same units as the observations. We report the mean and standard deviation of the normalized reward on 100 runs for 50 noise values between 0 and 2.}
    \label{fig:noise_indexing}
\end{figure}

Inferred indexing also proves to be a brittle solution.
To characterize this brittleness, we perform an evaluation by injecting observation noise during execution.
This is shown in \autoref{fig:noise_indexing}.
We report the mean and standard deviation of the normalized reward on 100 runs for 50 noise values between 0 and 2. As we can observe, all models start with the optimal policy with a reward of 1 when the noise is 0.
As the noise increases, we observe how homogeneous models either almost immediately lose functionality (like for Scenario B in \autoref{fig:noise_indexing_b}), or degrade in performance rapidly (like for Scenario A in \autoref{fig:noise_indexing_a}).
A heterogeneous policy, in contrast, is able to tolerate higher magnitudes of noise, and, even in the difficult corridor scenario, still manages to complete the task about 20\% of the time at high noise values.

%The practical effects of noise in either scenarios are very significant.
%Figure~\ref{fig:concepts-egA-egB} depicts this quantitatively by graphing the success of either policies under varying magnitudes of measurement noise.
%The ``reward" is redefined here such that it is normalized between 0 and 1.
%such that it is 1 when the task is completed (or optimal policies as the percentage of times the task is completed, i.e., the agents are able to get to their desired positions.
%We notice that in both cases, a homogeneous policy shows little robustness to noise -- when the measurements are perfect (noise=0), both homogeneous and heterogeneous policies perform similarly, however, the performance of a homogeneous policy degrades very rapidly.

%%%%%%%%%%%%%%%%%%%%%%%%%%%%%%%%%%%%%%%%%%%%%%%%%%%%%%%%%%%%%%%%%%%%%%%%%%%%%%%%
\section{Experimental evaluations}
\label{sec:evaluations}
We now present some evaluations of the proposed models in simulated and real environments. 

% Simulations are executed in custom created scenarios using the VMAS vectorized simulator~\cite{bettini2022vmas}. \fixit{either remove the before, or also add 1 sentence about real world, eg 'real world exp are performed in a x X xm lab with a robomaster robot, etc}

\begin{figure}[tb]
    \centering
    \begin{subfigure}{0.35\linewidth}
        \includegraphics[width=\linewidth]{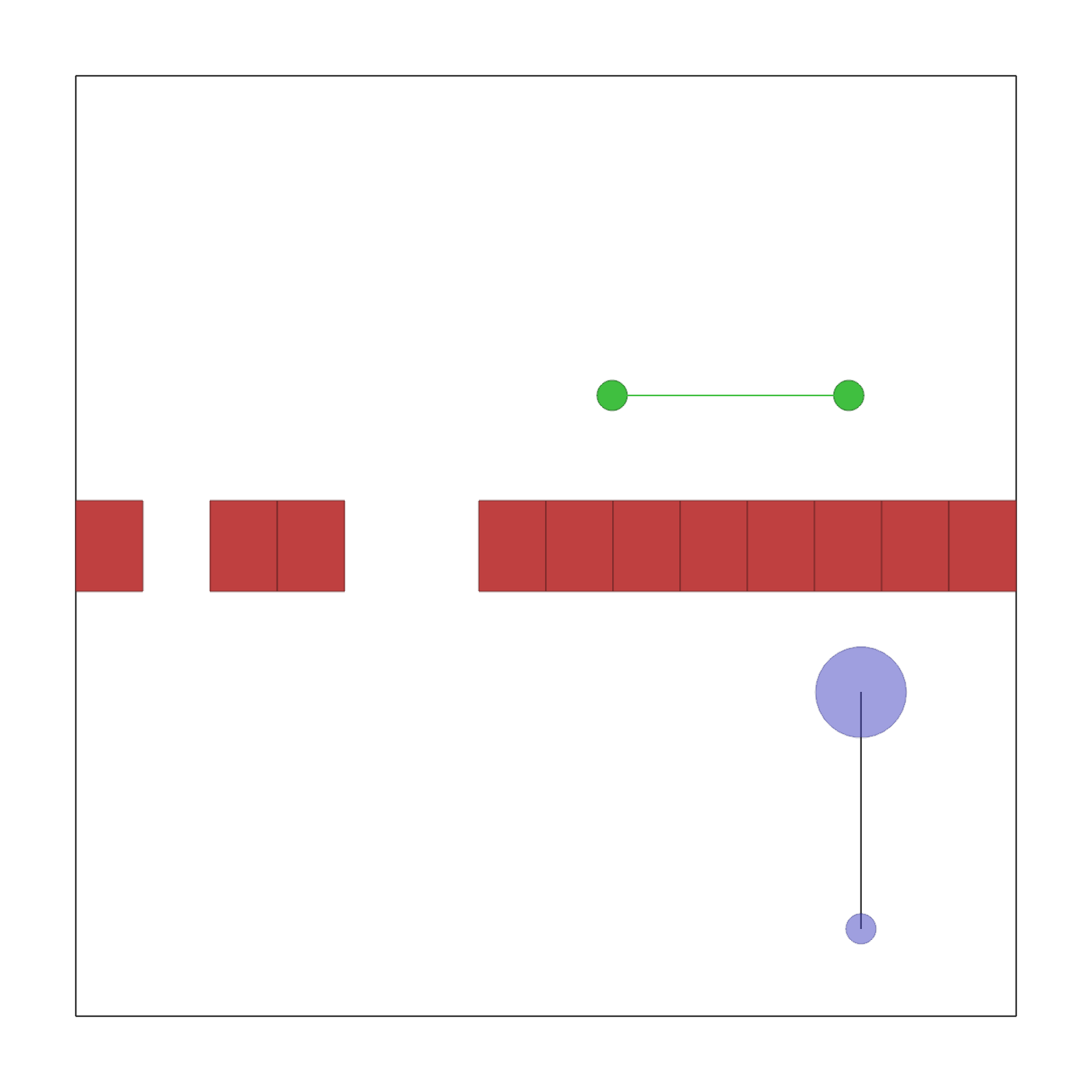}
        \vspace*{9pt}
        \caption{Scenario}
        \label{fig:joint_size_setup}
    \end{subfigure}%
    \begin{subfigure}{0.65\linewidth}
        \resizebox{\linewidth}{!}{%
        \input{figures/joint_size/joint_size_success.tex}}%
        \caption{Training performance}
        \label{fig:joint_size_success}
    \end{subfigure}
     \caption{Performance evaluation in the passage scenario with differently sized robots. Here, the homogeneous model is not able to perform \textit{inferred behavioral typing for physically heterogeneous agents} since $\mathcal{P}$ heterogeneity does not affect the robots' observations. Thus, only the heterogeneous model is able to solve the task. We plot the mean and standard deviation success rate of 4 runs. Each iteration is performed over 200 episodes of experience.}
     \label{fig:joint_size}
\end{figure}

\textbf{Performance evaluation}. 
We evaluate HetGPPO on a simulated 2D task which requires heterogeneous behavior. The task is shown in \autoref{fig:joint_size_setup}.
Here, two robots of different sizes (blue circles),
connected by a rigid linkage through two revolute joints,
need to cross a passage while keeping the linkage parallel to it and then match the desired goal position (green circles) on the other side.
The passage is comprised of a bigger and a smaller gap, which are spawned in a random position and order on the wall, but always at the same distance between each other.
The team is spawned in a random order and position on the lower side with the linkage always perpendicular to the passage.
The goal is spawned horizontally in a random position on the upper side. Each robot observes and communicates its velocity, relative position to each gap, and relative position to the goal center.  
The shaped global reward is composed of two convex terms. Before the passage, the robots are rewarded to keep the linkage parallel to the goal and to bring its center to the center of the passage. After the passage, the robots are rewarded for bringing it to the goal at the desired orientation. Collisions are also penalized.

\autoref{fig:joint_size_success} shows training success rate (i.e., percentage of episodes in each batch that complete the task).
The heterogeneous model is able to learn two behaviorally different policies: the bigger robot passes through the bigger gap and the smaller robot through the smaller gap, achieving the optimal solution.
On the other hand, the homogeneous model is not able to assign these two behavioral types using \textit{inferred behavioral typing for physically heterogeneous agents}, since the $\mathcal{P}$ heterogeneity caused by different robot sizes does not affect the robots' observations.
%Homogeneous agents
Agents with homogeneous policies never manage to cross the passage, being deterred by unavoidable collisions.
%The training success rate (i.e., percentage of episodes in each batch that complete the task) is reported in \autoref{fig:joint_size_success}.

\begin{figure}[tb]
    \centering
    \begin{subfigure}{0.35\linewidth}
        \includegraphics[width=\linewidth]{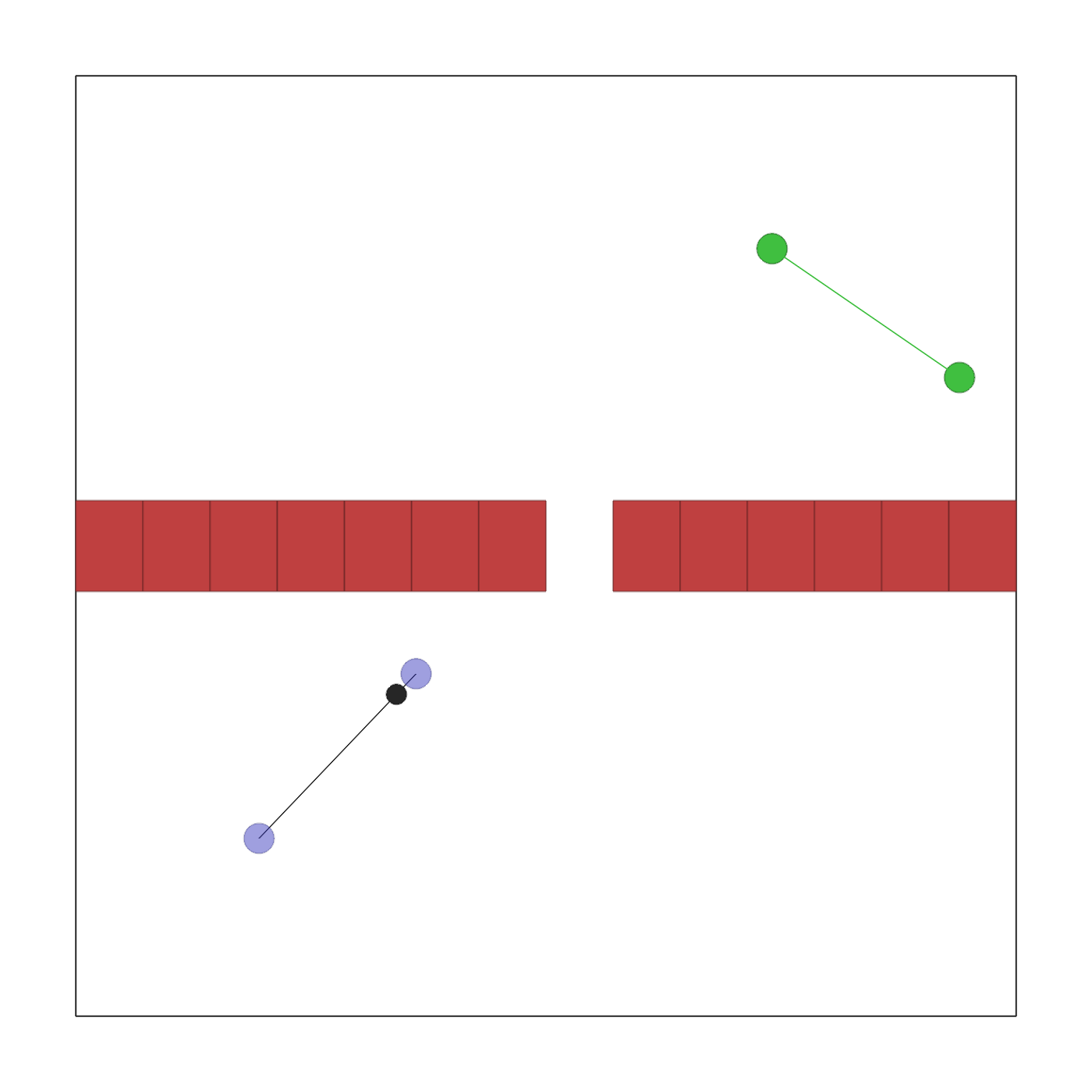}
        \vspace*{4pt}
        \caption{Scenario}
        \label{fig:training_noise_scenario}
    \end{subfigure}%
    \begin{subfigure}{0.65\linewidth}
        \includegraphics[width=\linewidth]{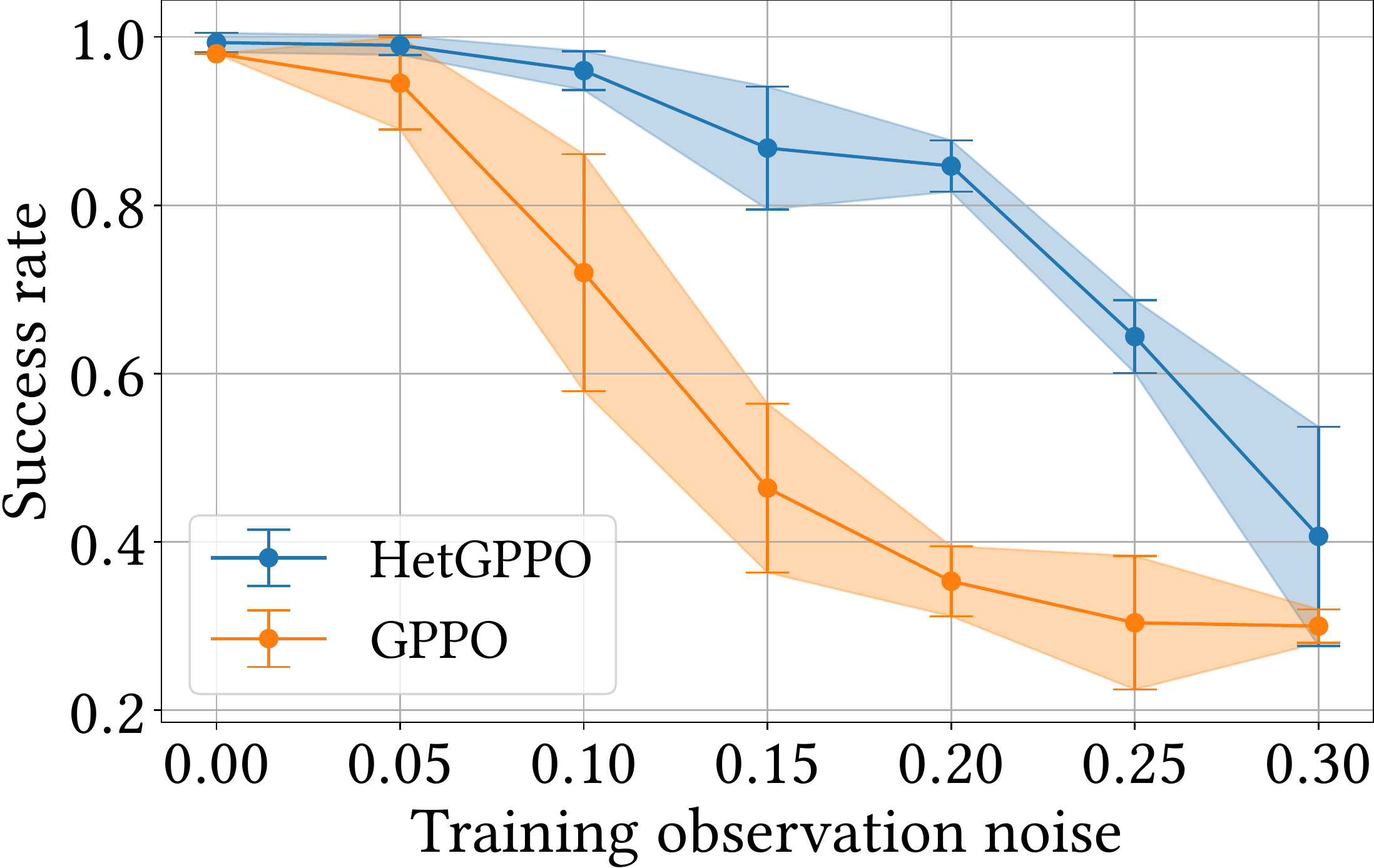}
        \caption{Resilience to training noise}
        \label{fig:training_noise_eval}
    \end{subfigure}
     \caption{Resilience to uniform observation noise during training in the passage scenario with asymmetric package. Here, the heterogeneous model is able to  maintain higher performance as the noise increases. We train the two models for 7 different noise values. For each noise value, we report the mean and standard deviation of the success rate after 1000 training iterations for 5 runs. Each training iteration is performed over 200 episodes of experience.}
    \label{fig:training_noise}
\end{figure}

\textbf{Resilience to training noise}. 
As elucidated in \autoref{sec:typing}, homogeneous models can learn heterogeneous behavior. In this subsection, we evaluate the resilience of this paradigm in the presence of observation noise during training.
We consider the task depicted in \autoref{fig:training_noise_scenario}.
This is the same as in \autoref{fig:joint_size_setup} with the difference that the robots are now physically identical, but the linkage has an asymmetric mass (black circle) that causes a different type of $\mathcal{P}$ heterogeneity, reflected in the velocity observations.
The passage is a single gap, positioned randomly on the wall. The agents need to cross it while keeping the linkage perpendicular to the wall and avoiding collisions. The team and the goal are spawned in a random position, order, and rotation on opposite sides of the passage.

In \autoref{fig:training_noise_eval} we report the training success rate for different observation noise values.
Thanks to \textit{inferred behavioral typing for physically heterogeneous agents} we see that both models solve the task optimally when $0$ noise is added.
As noise increases, the heterogeneous model is able to maintain significantly better performance. For example, with 0.2 observation noise, HetGPPO still achieves more than 80\% success rate, while GPPO is below 40\%.
 
\begin{figure}[tb]
    \centering
    \begin{subfigure}{0.253\linewidth}
    \includegraphics[width=\linewidth]{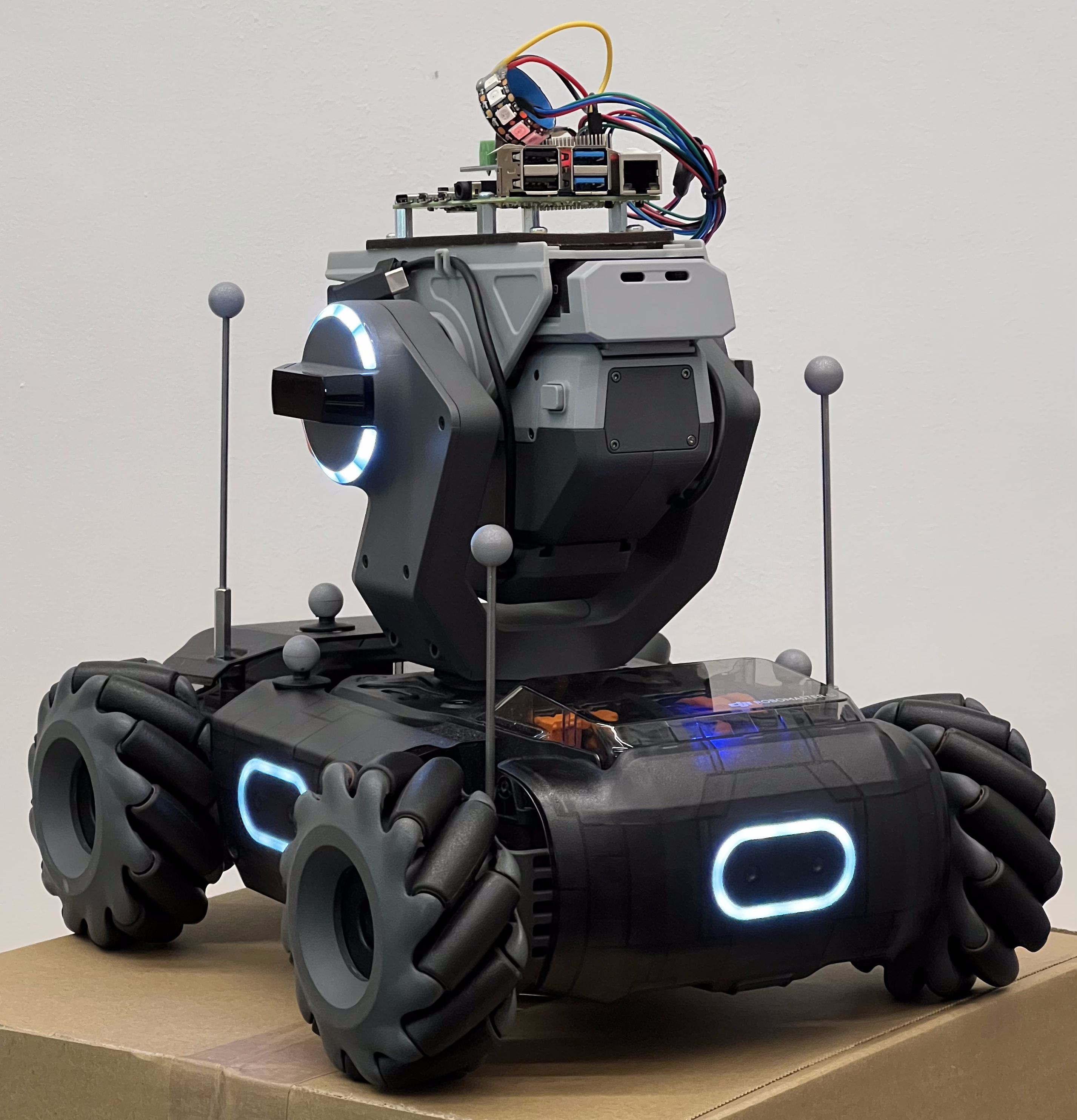}
    \caption{Robot}
    \label{fig:real_world_robot}
    \end{subfigure}
    \hfill
    \begin{subfigure}{0.655\linewidth}
    \includegraphics[width=\linewidth]{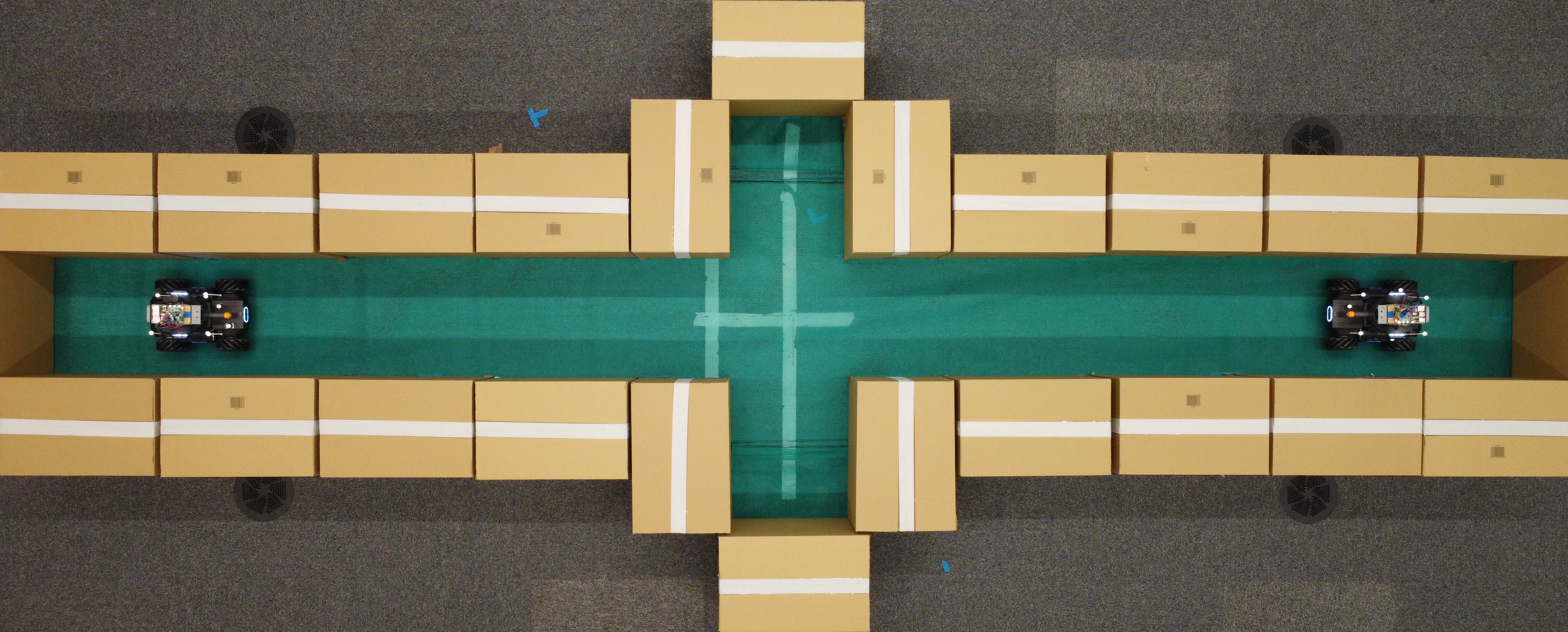}
    \caption{Scenario B (real world)}
    \label{fig:real_world_setup}
    \end{subfigure}
    \par\medskip
    \begin{subfigure}{0.48\linewidth}
        \includegraphics[width=\linewidth]{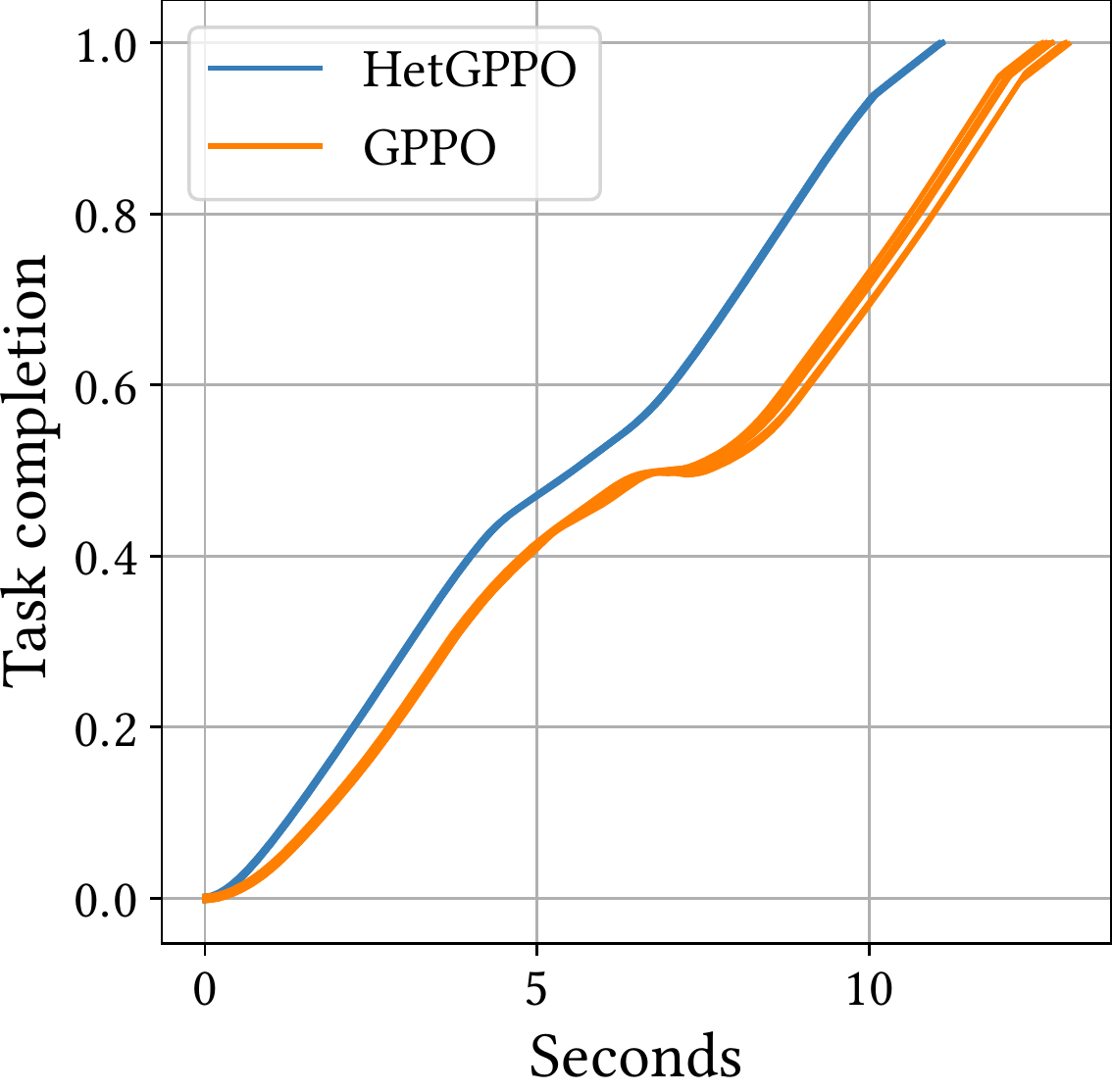}
        \caption{Simulation}
        \label{fig:deployment_sim}
    \end{subfigure}
    \hfill
    \begin{subfigure}{0.48\linewidth}
        \includegraphics[width=\linewidth]{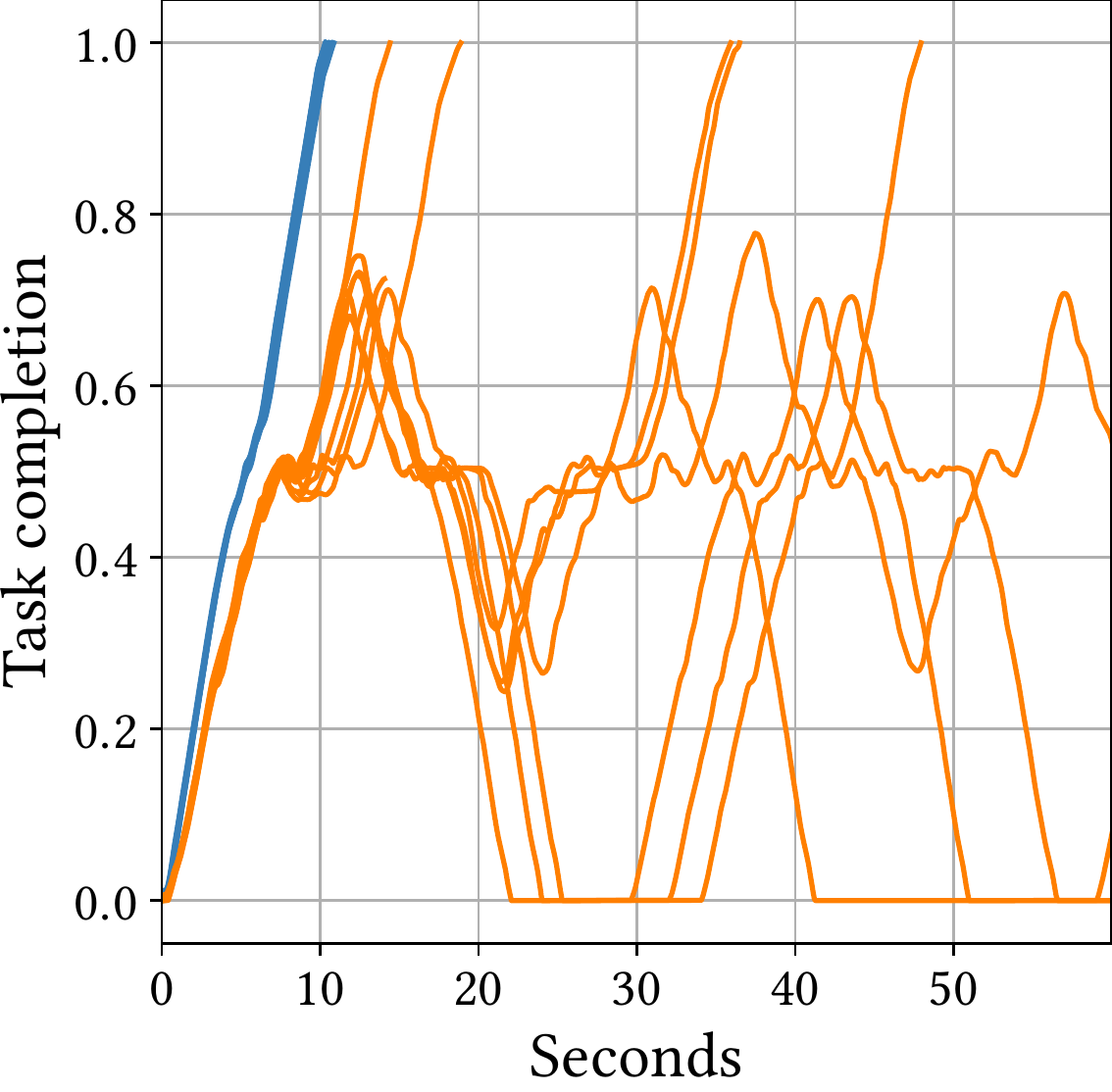}
        \caption{Real world}
        \label{fig:deployment_real}
    \end{subfigure}
    \caption{Real-world deployment of Scenario B (\autoref{fig:scenario_b}). We report 10 runs for each model both in simulation and in the real world. We plot task completion (the scaled sum of the negative distances of each robot from its goal) over time. While in simulation both models are able to perform the task, real world imperfections make the homogeneous model dynamically switch between learned behavioral types, leading to the robots switching positions multiple times near the central area. This causes the zigzag behavior in (d) with certain rollouts failing or taking over the maximum allocated time of 60s.}
    \label{fig:deployment}
\end{figure}

\textbf{Real-world deployment}.
To demonstrate the resilience of heterogeneous policies, we deploy Scenario B (\autoref{sec:inferred_typing_identical}) to a real-world setting.
The setup of the task is shown in \autoref{fig:real_world_setup} and is the same as in simulation.
We use two holonomic RoboMaster S1 ground robots~\cite{Robomaster2}
%\footnote{\url{https://www.dji.com/uk/robomaster-s1}}
(\autoref{fig:real_world_robot}), each running a customized model-based controller onboard~\cite{shankar2021freyja}.
We perform 10 runs for the trained HetGPPO and GPPO models both in simulation (\autoref{fig:deployment_sim}) and in the real world (\autoref{fig:deployment_real}).
As already discussed in \autoref{sec:inferred_typing_identical}, both the heterogeneous and the homogeneous models are able to solve the scenario in simulation, with the homogeneous model leveraging \textit{inferred behavioral typing for physically identical agents}.
This is shown in \autoref{fig:deployment_sim}, where all the runs of both models reach 100\% task completion within 15s.
On the other hand, as seen in \autoref{fig:deployment_real}, the performance of the homogeneous model is heavily impacted in the real world.
This is because, in this symmetric scenario, the homogeneous model cannot type agents based on position only, and has to rely on velocity observations to build the behavioral types.
%Thus, due to imperfections in the robots' velocities at certain timesteps, the model cannot distinguish if the robots are currently moving towards the center or away from it and erroneously dynamically switches the behavioral types of the two robots.
%Since these velocities are often noisy, the model fails to distinguish if the robots are currently moving towards or away from the center (since it has no memory), and dynamically switches (erroneously) the behavioral types of the two robots.
%Such velocity noise is caused by the interaction between the neural network and the onboard controller. This, alongside control delays, causes imperfections unseen in simulation and cannot be easily fixed (e.g., by filtering).
In practice, however, real-world estimated velocities can be noisy due to factors such as control/process delays and variability in the robot's measurement and motion models.
Thus, relying on these observations makes the homogeneous (memory-less) model susceptible to erroneously switching the behavioral types dynamically (i.e., failing to distinguish if the robots are currently moving towards or away from the center).
This leads to the plotted rollouts, where robots alternate the role of giving way to each other near the passage. Out of 10 runs, only 5 are completed within 60s. The heterogeneous model, on the other hand, does not rely on behavioral typing and is not impacted by the deployment noises, performing as well as in simulations.

%\begin{figure}[ht]
%    \centering\hfill
%    \begin{subfigure}{0.5\linewidth}
%       \resizebox{\linewidth}{!}{%
%       \input{figures/real_world/simualted/trial.tex}}
%       \caption{Simulation}
%    \end{subfigure}\hfill
%    \begin{subfigure}{0.5\linewidth}
%        \resizebox{\linewidth}{!}{%
%       \input{figures/real_world/simualted/trial.tex}}
%        \caption{Real-world}
%    \end{subfigure}\hfill
%    \caption{P.}
%\end{figure}

%%%%%%%%%%%%%%%%%%%%%%%%%%%%%%%%%%%%%%%%%%%%%%%%%%%%%%%%%%%%%%%%%%%%%%%%
\section{Conclusion}
\label{sec:conclusion}

In this paper, we introduced a new paradigm for learning heterogeneous policies in MARL. We motivated it with a categorization of techniques that homogeneous models can use to emulate heterogeneous behavior and empirically demonstrated their limits. Finally, we showed the benefits of policy heterogeneity for both performance and resilience on multi-robot tasks in simulation and in the real world.
While we do not employ any methods to control the degree of heterogeneity of the agents’ policies, we observe that training is already a good heterogeneity regularizer. In other words, if the system has heterogeneous requirements, HetGPPO will be able to learn them, while, if the system benefits from homogeneous policies, HetGPPO will learn the same policy as GPPO (with some loss in sample efficiency). 
In future work, we are interested in developing mechanisms that measure and actively tune the degree of policy heterogeneity in the team, allowing us to control the trade-offs between sample efficiency (of homogeneous policies) and resilience (of heterogeneous policies). %Further investigations will be made into the unexplored benefits of `apparently heterogeneous' homogeneous policies.
%Furthermore, there might be additional unexplored benefits for teams lying at the homogeneous end of this trade-off. 
%For instance, behavioral typing could be used to dynamically switch agent behavioral roles throughout execution.

%%%%%%%%%%%%%%%%%%%%%%%%%%%%%%%%%%%%%%%%%%%%%%%%%%%%%%%%%%%%%%%%%%%%%%%%

\section*{Acknowledgments}
This work was supported by ARL DCIST CRA W911NF-17-2-0181, the European Research Council (ERC) Project 949940 (gAIa), and in part by a gift from Arm.

%%% The next two lines define, first, the bibliography style to be 
%%% applied, and, second, the bibliography file to be used.

\bibliographystyle{ACM-Reference-Format} 
\bibliography{bibliography}

%%%%%%%%%%%%%%%%%%%%%%%%%%%%%%%%%%%%%%%%%%%%%%%%%%%%%%%%%%%%%%%%%%%%%%%%%%%%%%%%

\appendix
\setcounter{section}{0}
%%%%%%%%%%%%%%%%%%%%%%%%%%%%%%%%%%%%%%%%%%%%%%%%%%%%%%%%%%%%%%%%%%%%%%%%%%%%%%%%
\section{Experimental setup}

\subsection{Simulation}
We attach all the code used for simulations and training.
Simulations are performed in the VMAS~\cite{bettini2022vmas} simulator. All environments are customly created apart from Scenario B which is adapted from one of the scenarios already available in the simulator.  The training is performed in RLlib~\cite{liang2018rllib} using PyTorch~\cite{paszke2019pytorch} and an implementation of the PPO algorithm for multi-agent training. The general training parameters used are shown in \autoref{tab:rl_paramz}. Small variations of these are done on a per-environment basis and can be seen in the training scripts attached. The GPPO and HetGPPO model implementations and details are available in the code. Training is performed on a NVIDIA GeForce RTX 2080 Ti GPU. Each worker collects experience from the simulator using an Intel(R) Xeon(R) Gold 6248R CPU @ 3.00GHz. 

\begin{table}[h]
    \centering
    \caption{PPO training parameters.}
    \label{tab:rl_paramz}
    \begin{tabular}{lc|cc}
    \multicolumn{2}{c}{Training} & \multicolumn{2}{c}{PPO} \\ \toprule
    Batch size & 60000  & $\epsilon$ & 0.2 \\ 
    Minibatch size & 4096 & $\gamma$ & 0.99  \\ 
    SDG Iterations & 40 & $\lambda$ & 0.9 \\ 
    \# Workers &  5 & Entropy coeff & 0 \\
    \# Envs per worker &  50 & KL coeff & 0.01 \\
    Learning rate & 5e-5 & KL target & 0.01\\
    \end{tabular}
\end{table}

\subsection{Real-world}
Real-world experiments are performed using an Optitrack\footnote{\url{https://optitrack.com/}} motion capture system with 12 cameras to provide positional information to the robots. The robots used are holonomic RoboMaster S1 ground robots\footnote{\url{https://www.dji.com/uk/robomaster-s1}}, running a customized model-based controller onboard~\cite{shankar2021freyja}.
%%%%%%%%%%%%%%%%%%%%%%%%%%%%%%%%%%%%%%%%%%%%%%%%%%%%%%%%%%%%%%%%%%%%%%%%%%%%%%%%
\section{Scenario b reward structure}

The reward used to train Scenario B is comprised of two components: a positional reward and a collision reward. 

The positional reward is proportional to the time delta in relative distance of an agent from its goal. In other words, a positive reward is assigned if an agent moves towards its goal and a negative one if it moves away. The agents receive a shared positional reward equal to the sum of their individual positional rewards. When both agents are placed on their goal, they keep receiving an additional final reward. The episode ends after 500 timesteps.

The collision reward is a constant penalty assigned to each agent in the presence of collisions. When training starts, the only collisions penalized are inter-agent ones. A curriculum is set up throughout training so that, when the agents' positional reward gets high enough to symbolize that they solved the task, collisions at the recesses start being penalized as well. This is done so that the agents are able to first learn to solve the task and can then fine-tune their performance by removing collisions.

%%%%%%%%%%%%%%%%%%%%%%%%%%%%%%%%%%%%%%%%%%%%%%%%%%%%%%%%%%%%%%%%%%%%%%%%%%%%%%%%
\section{Sim to real transfer}

To deploy policies trained in the VMAS simulator to the real world, we iteratively tune some simulation hyperparameters to fit the real-world conditions. These parameters are dependent just on the robots and their interaction with the real-world. Once tuned, they can be used for any training scenario. 

The parameters that were key to a successfully deployment are linear friction and drag. Since we operate with ground robots at relatively low speeds, we set drag to 0 and tune linear friction. Through 3 real to sim iterations of binary search we were able to find the correct friction value for our robot-ground pair. Together with friction, we tuned the maximum acceleration in simulation to fit the real robot one. These parameters were tested and validated on simple single-robot tasks such as trajectory following and moving to a goal position.

%%%%%%%%%%%%%%%%%%%%%%%%%%%%%%%%%%%%%%%%%%%%%%%%%%%%%%%%%%%%%%%%%%%%%%%%

\end{document}